\documentclass{article}
\usepackage{ijcai22}

\usepackage{times}
\usepackage{soul}
\usepackage{url}
\usepackage{xr-hyper}
\usepackage[hidelinks]{hyperref}
\usepackage[utf8]{inputenc}
\usepackage[small]{caption}
\usepackage{graphicx}
\usepackage{booktabs}
\usepackage{algorithm}
\usepackage{algorithmic}
\usepackage{latexsym}
\usepackage{amssymb,amsmath,amsthm}
\usepackage{multirow}
\usepackage{multicol}

\usepackage{xr}
\makeatletter
\newcommand*{\addFileDependency}[1]{
  \typeout{(#1)}
  \@addtofilelist{#1}
  \IfFileExists{#1}{}{\typeout{No file #1.}}
}
\makeatother

\newcommand*{\myexternaldocument}[1]{%
    \externaldocument{#1}%
    \addFileDependency{#1.tex}%
    \addFileDependency{#1.aux}%
}
\myexternaldocument{supp}

\usepackage[algo2e]{algorithm2e}

\newcommand{\xv}{{\boldsymbol x}}
\newcommand{\yv}{{\boldsymbol y}}
\newcommand{\Xv}{{\boldsymbol X}}
\newcommand{\Yv}{{\boldsymbol Y}}
\newcommand{\wv}{{\boldsymbol w}}

\title{AdaK-NER: An Adaptive Top-$\mathbf{K}$ Approach \\
for Named Entity Recognition with Incomplete Annotations}

\author{
Hongtao Ruan$^1$\and
Liying Zheng$^2$\And
Peixian Hu$^{3}$
}

\begin{document}

\maketitle

\begin{abstract}
State-of-the-art Named Entity Recognition (NER) models rely heavily on large amounts of fully annotated training data. 
However, accessible data are often incompletely annotated in the industry like Financial Technology.
Normally the unannotated tokens are regarded as non-entities by default, while we underline that these tokens could either be non-entities or part of any entity. Here, we study NER modeling with incomplete annotated data where only a fraction of the named entities are labeled, and the unlabeled tokens are equivalently multi-labeled by every possible label. Taking multi-labeled tokens into account, the numerous possible paths can distract the training model from the gold path (ground truth label sequence), and thus hinders the learning ability. In this paper, we propose AdaK-NER, named the adaptive top-$K$ approach, to help the model focus on a smaller feasible region where the gold path is more likely to be located. We demonstrate the superiority of our approach through extensive experiments on both English and Chinese datasets, averagely improving $2\%$ in F-score on the CoNLL-2003 and over $10\%$ on two Chinese datasets from Financial Technology compared with the prior state-of-the-art works.
\end{abstract}

\section{Introduction}

Named Entity Recognition (NER) \cite{li2020flat,sang2003introduction,peng2019distantly} is a fundamental task in Natural Language Processing (NLP). NER task aims at recognizing the meaningful entities occurring in the text, which can benefit various downstream tasks, such as question answering \cite{cao2019bag}, event extraction \cite{wei2020study}, and opinion mining \cite{poria2016aspect}. Especially, by detecting relevant entities in Financial Technology, NER helps financial professionals efficiently leverage the information of news, which is paramount in making high-quality decisions.

Strides in statistical models, such as Conditional Random Field (CRF) \cite{lafferty2001conditional} and pre-trained language models like BERT \cite{devlin2018bert}, have equipped NER with new learning principles \cite{li2020flat}. 
Pre-trained model with rich representation ability
can discover hidden features automatically while CRF can capture the dependencies between labels with the BIO or BIOES tagging scheme. 

\begin{figure*}[!htbp]
\centering
\includegraphics[width=.85\textwidth]{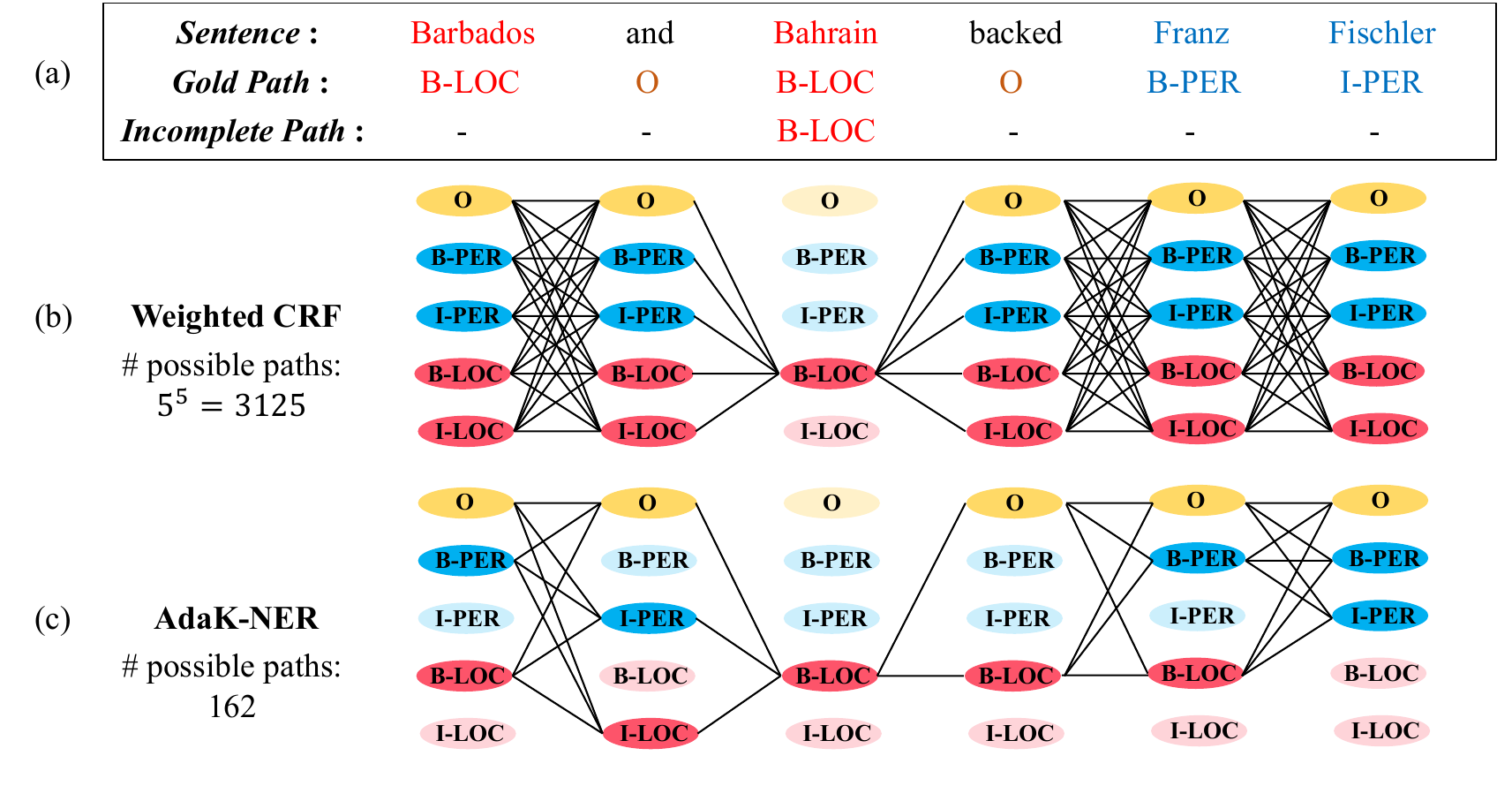}
\caption{(a) The sentence is annotated with BIO tagging scheme. The entity types are person (PER) and location (LOC). Only the entity `Bahrain' of LOC is recognized, while `Barbados' of LOC and `Franz Fischler' of PER are missing. (b) Weighted CRF model considers all the $3125$ possible paths with $5$ unannotated tokens for $q$ estimation. (c) In our model, we build a candidate mask to filter out the less likely labels (labels in faded color). Therefore, the possible paths of our model is significantly less than weighted CRF.}
\label{crf_paths}
\end{figure*}

However, most existing methods rely on large amounts of fully annotated information for training NER models \cite{li2020flat,jia2020entity}.
Fulfilling such requirements is expensive and laborious in the industry like Financial Technology. Annotators, are not likely to be fully equipped with comprehensive domain knowledge, only annotate the named entities they recognize and let the others off, resulting in incomplete annotations. They typically do not specify the non-entity \cite{surdeanu2010legal}, so that the recognized entities are the only available annotations. Figure \ref{crf_paths}(a) shows examples of both gold path\footnote{A path is a label sequence for a given sentence.} and incomplete path. 

For corpus with incomplete annotations, each unannotated token can either be part of an entity or non-entity, making the token equivalently multi-labeled by every possible label. Since conventional CRF algorithms require fully annotated sentences, a strand of literature suggests assigning weights to possible labels \cite{shang2018learning,jie2019better}. Fuzzy CRF \cite{shang2018learning} focused on filling the unannotated tokens by assigning equal probability to every possible path. Further, Jie~\shortcite{jie2019better} introduced a weighted CRF method to seek a more reasonable distribution $q$ for all possible paths, attempting to pay more attention to those paths with high potential to be gold path.

Ideally, through comprehensive learning on $q$ distribution, the gold path can be correctly discovered.
However, this perfect situation is difficult to achieve in practical applications. 
Intuitively, taking all possible paths into consideration will distract the model from the gold path, as the feasible region (the set of possible paths where we search for the gold path) grows exponentially with the length of the unannotated tokens increasing, which might cause failure to identify the gold path.

To address this issue, one promising direction is to prune the size of feasible region during training. We assume the unknown gold path is among or very close to the top-$K$ paths with the highest possibilities.
Specifically, we utilize a novel adaptive $K$-best loss to help the training model focus on a smaller feasible region where the gold path is likely to be located.
Furthermore, once a path is identified as a disqualified sequence, it will be removed from the feasible region. This operation can thus drastically eliminate redundancy without undermining the effectiveness. For this purpose, a candidate mask is built to filter out the less likely paths, so as to restrict the size of the feasible region.


Trained in this way, our AdaK-NER overcomes the shortcomings of Fuzzy CRF and weighted CRF, resulting in a significant improvement on both precision and recall, and a higher $F_1$ score as well.

In summary, the contributions of this work are:
\vspace{-.2em}
\begin{itemize}
    \item We present a $K$-best mechanism for improving incomplete annotated NER, aiming to focus on the gold path effectively from the most possible candidates.
    \vspace{-.3em}
    \item We demonstrate both qualitatively and quantitatively that our approach achieves state-of-the-art performance compared to various baselines on both English and Chinese datasets.
\end{itemize}

\vspace{-8pt}
\section{Proposed Approach}
\vspace{-1pt}

Let $\xv=(\xv_1, \xv_2, \cdots, \xv_n)$ be a training sentence of length $n$, token $\xv_i\in \Xv$. Correspondingly, $\yv=(\yv_1, \yv_2,\cdots, \yv_n)$ denotes the complete label sequence, $\yv_i \in \Yv$. 
The NER problem can be defined as inferring $\yv$ based on $\xv$.

Under the incomplete annotation framework, we introduce the following terminologies. A possible path refers to a possible complete label sequence consistent with the annotated tokens. 
For example, a possible incomplete annotated label sequence for $\xv$ can be $\yv_u = (-,\yv_2,-, \cdots, -)$, where token $\xv_2$ is annotated as $\yv_2$ and other missing labels are labeled as $-$. $\yv_c = (\yv_{c_1}, \yv_2, \cdots, \yv_{c_n})$ with $\yv_{c_i}\in \Yv$, is a possible path for $\xv$, where all the missing labels $-$ are replaced by some elements in $\Yv$. Set $C(\yv_u)$ denotes all possible complete paths for $\xv$ with incomplete annotation $\yv_{u}$. $D=\{(\xv^i,\yv^i_u)\}$ is the available incompletely annotated dataset.

For NER task, CRF \cite{lafferty2001conditional} is a traditional approach to capture the dependencies between the labels by modeling the conditional probability $p_\wv(\yv|\xv)$ of a label sequence $\yv$ given an input sequence $\xv$ of length $n$ as:
\hspace{-1em}
\begin{equation}
    p_\wv(\yv|\xv)=\frac{\exp(\wv \cdot \Phi(\xv,
    \yv))}{\sum_{\yv \in \Yv^n} \exp(\wv\cdot \Phi(\xv,\yv))}.
\end{equation}
$\Phi(\xv,\yv)$ denotes the map from a pair of $\xv$ and $\yv$ to an arbitrary feature vector, $\wv$ is the model parameter, $p_\wv(\yv|\xv)$ is the probability of $\yv$ predicted by the model. Once $\wv$ has been estimated via minimizing negative log-likelihood:
\begin{equation}\label{loss L}
    L(\wv,\xv) = -\log p_\wv(\yv|\xv),
\end{equation}
the label sequence can be inferred by:
\begin{equation}\label{viterbi}
    \hat{\yv} =\arg\max_{\yv \in \Yv^n} p_\wv(\yv |\xv).
    \vspace{-1pt}
\end{equation}

The original CRF learning algorithm requires a fully annotated sequence $\yv$, thus the incompletely annotated data is not directly applicable to it. Jie~\shortcite{jie2019better} modified the loss function as follows:
\begin{equation}\label{loss weighted}
    L(\wv, \xv) = -\log \sum_{\yv\in C(\yv_u)}q(\yv|\xv)p_\wv(\yv|\xv),
    \vspace{-1pt}
\end{equation}
where $q(\yv|\xv)$ is an estimated distribution of all possible complete paths $\yv \in C(\yv_u)$ for $\xv$. 

We illustrate their model in (Figure \ref{crf_paths}b). Note that when $q$ is a uniform distribution, the above CRF model is Fuzzy CRF \cite{shang2018learning} which puts equal probability to all possible paths in $C(\yv_u)$. Jie~\shortcite{jie2019better} claimed that $q$ should be highly skewed rather than uniformly distributed, therefore they presented a $k$-fold cross-validation stacking method to approximate distribution $q$.

Nonetheless, as Figure \ref{crf_paths}(b) shows, a sentence with only $6$ words ($1$ annotated, $5$ unannotated) have $3125$ possible paths. 
We argued that identifying the gold path from all possible paths is like looking for a needle in a haystack.
This motivates us to reduce redundant paths during training. 
We propose two major strategies (adaptive $K$-best loss and candidate mask) to induce the model to focus on the gold path (Figure \ref{crf_paths}(c)), and two minor strategies (annealing technique and iterative sample selection) to further improve the model effectiveness in NER task. 
The workflow is summarized in Algorithm \ref{detail-alg}.

\vspace{-3pt}
\subsection{Adaptive $K$-best Loss}\label{section_4_2}
Viterbi decoding algorithm is a dynamic programming technique to find the most possible path with only linear complexity, thus it could be used to predict a path for an input based on the parameters provided by the NER model.
$K$-best Viterbi decoding \cite{huang2005better} extends the original Viterbi decoding algorithm to extract the top-$K$ paths with the highest probabilities. In the incomplete data, the gold path is unknown. We hypothesize it is very likely to be the same with or close to one of the top-$K$ paths. 
This inspires us to introduce an auxiliary $K$-best loss component to help the model focus on a smaller yet promising region. 
Weight is added to balance the weighted CRF loss and the auxiliary loss, and thus we modify \eqref{loss weighted} into:
\vspace{-1pt}
\begin{equation}\label{K-best loss}
\begin{aligned}
    L_k(\wv,\xv) =& -(1-\lambda)\log\sum_{\yv\in C(\yv_u)}q(\yv|\xv)p_\wv(\yv|\xv)\\ 
    &- \lambda \log \sum_{\yv\in K_\wv(x)}p_w(\yv|\xv),
\end{aligned}
\vspace{-1pt}
\end{equation}
where $K_\wv(\xv)$ represents the top-$K$ paths decoded by constrained $K$-best Viterbi algorithm\footnote{The constrained decoding ensures the resulting complete paths are compatible with the incomplete annotations.} with parameters $\wv$, and $\lambda$ is an adaptive weight coefficient.


\vspace{-3pt}
\subsection{Estimating $q$ with Candidate Mask}\label{section_4_1}
We divide the training data into $k$ folds and employ $k$-fold cross-validation stacking to estimate $q$ for each hold-out fold \cite{jie2019better}.
We propose an interpolation mode to adjust $q$ by increasing the probabilities for paths with high confidence and decreasing for the others.
The probability of each possible path is a temperature softmax of $\log p_{\wv_i}$:
\begin{equation}\label{temperature}
\begin{aligned}
   q_{\wv_i}(\yv|\xv)=
    \frac{\exp\left(\log p_{\wv_i}(\yv|\xv)/T\right)}
    {\sum_{\yv}
    \exp \left(\log p_{\wv_i}(\yv|\xv)/T\right)},
\end{aligned}
\end{equation}
where $T>0$ denotes the temperature and $\wv_i$ is the model trained by holding out the $i$-th fold. 
A higher temperature produces a softer probability distribution over the paths, resulting in more diversity and also more mistakes \cite{hinton2015distilling}. 
We iterate the cross-validation until $q$ converges.
Jie~\shortcite{jie2019better} estimated $q(\yv|\xv)$ for each $\yv\in C(\yv_u)$ while the size of $C(\yv_u)$ grows exponentially on the number of unannotated tokens in $\xv$.
To reduce the number of possible paths for $q$ estimation, we build candidate mask based on the $K$-best candidates and the self-built candidates.

\begin{algorithm}[H]
 \textbf{Data}: $D=\{(\xv^i, \yv_u^i)\}$ \\
 Randomly divide $D$ into $k$ folds: $D_1, D_2, \cdots, D_k$ \\
 Entity Dictionary $\mathcal{H}\leftarrow \emptyset$ \\
 Initialize model $M$ with parameters $\hat{\wv}$ \\
 Initialize $q$ distributions \{$q(\cdot|\xv^i)$\} \\
 Sample importance score $s_i\leftarrow 1$ \\
 hyper-parameters $s$ and $c$ \\
 \For{iteration = $1,\cdots, N$}{
 \% Sample Selection \\
  $D^{'}\leftarrow D$ \\
  \For{j = $1,\cdots, k$}{
  $D_{j}^{'}\leftarrow D_{j}$ \\
  \For{$(\xv^i, \yv_u^i)\in D_{j}^{'}$}{
  \If{$s_i < s$}{remove $(\xv^i, \yv_u^i)$ from $D_{j}^{'}$ and $D^{'}$}
  }}
  \% $q$ Distribution Estimating \\
  \For{j = $1,\cdots, k$}{
  Train $M(\wv_j)$ on $D^{'}\backslash D_{j}^{'}$: Eq.(7)\\
  \For{$(\xv^i, \yv_u^i)\in D_{j}^{'}$}{
  Predict $K_b(\xv^i)$ by $M(\hat{\wv})$ \\
  Extract $H(\xv^i)$ by $\mathcal{H}$ \\
  Possible paths {\small$S = S(\yv_u^i, K_b(\xv^i), H(\xv^i))$} \\
  Estimate $q(\yv|\xv^i)$ for $\yv \in S$: Eq.(6) \\
  $s_i = \max_\yv p_{\wv_j}(\yv|\xv^i)$ \\
  $e_i \leftarrow $ \{$entities$\} predicted by $M(\wv_j)$
  }}
  \% Dictionary $\mathcal{H}$ Updating \\
  $\mathcal{H}\leftarrow \emptyset$ \\
 \For{$entity \in \cup e_i$}{
 \If{$entity \notin \mathcal{H}$ and $freq(entity)$ \textgreater $c$}{
 $\mathcal{H} \leftarrow add\_entity(\mathcal{H}, entity)$}}
 Train $M(\wv^\prime)$ on $D$ with $q$: Eq.(7)\\
 \If{$F_1$ of $M(\wv^\prime) > F_1$ of $M(\hat{\wv})$ on Dev}{
 $\hat{\wv} \leftarrow \wv^\prime$
 }}
 Return the final NER model $M(\hat{\wv})$
 \caption{AdaK-NER} \label{detail-alg}
\end{algorithm}

\paragraph{$K$-best Candidates.}
During the end of each iteration, we train a model $M(\hat{\wv})$ on the whole training data $D$. In the next iteration, we use the trained model $M(\hat{\wv})$ to identify $K$-best candidates set $K_b(\xv)$ for each sample $\xv$ by constrained $K$-best Viterbi decoding. $K_b(\xv)=\{\hat{K}_i(\xv)\}_{i=1,\cdots,K}$ contains top-$K$ possible paths with the highest probabilities, where $\hat{K}_i(\xv) = [\hat{K}_i(\xv_1), \hat{K}_i(\xv_2), \cdots, \hat{K}_i(\xv_n)]$.

\paragraph{Self-built Candidates.}
In the current iteration, after training a model $M(\wv_i)$ on $(k-1)$ folds, we use $M(\wv_i)$ to predict a path for each sample in the hold-out fold, and extract entities from the predicted path. Then we merge all entities identified by $k$ models $\{M(\wv_i)\}_{i=1,\cdots,k}$,
resulting an entity dictionary $\mathcal{H}$. For each sample $\xv$ we conjecture that its named entities should lie in the dictionary $\mathcal{H}$. Consequently, in the next iteration we form a self-built candidate $H(\xv) = [H(\xv_1), H(\xv_2), \cdots, H(\xv_n)]$ for each $\xv$ of length $n$. $H(\xv_j)$ is the corresponding entity label if the token $\xv_j$ is part of an entity in $\mathcal{H}$, otherwise $H(\xv_j)$ is O label.

We utilize the above candidates ({\it i.e.}, the $K$-best candidates set $K_b(\xv)$ and the self-built candidate $H(\xv)$) to construct a candidate mask for $\xv$. 
For each unannotated $\xv_j$ in $\xv$, the possible label set consists of (1) O label (2) $H(\xv_j)$, (3) $\cup_{i=1}^K \hat{K}_i (\xv_j)$.

For example, as Figure \ref{crf_paths}(c) shows, the unannotated token `Barbados' is predicted as B-PER and B-LOC in the above candidate paths, we treat B-PER, B-LOC and O label as the possible labels of `Barbados' and mask the other labels. 

With this masking scheme, we can significantly narrow down the feasible region of $\xv$ (Figure \ref{crf_paths}(c)) when estimating $q(\cdot|\xv)$.
After estimating $q(\cdot|\xv)$, we can train a model through the modified loss:
\vspace{-3pt}
\begin{equation}\label{k-best masking loss}
\begin{aligned}
    L_{m}(\wv,\xv) = & -(1-\lambda)\log\sum_{\yv\in S}q(\yv|\xv)p_\wv(\yv|\xv)\\ 
    &- \lambda \log \sum_{\yv\in K_\wv(\xv)}p_\wv(\yv|\xv),
\end{aligned}
\vspace{-1pt}
\end{equation}
where $S = S(\yv_u, K_b(\xv), H(\xv))$ contains the possible paths restricted by the candidate mask.


\begin{table}
\centering
\scalebox{0.8}{\begin{tabular}{lcccccc}
\hline 
\multirow{2}*{\textbf{Dataset}} & \multicolumn{2}{c}{\textbf{Train}} & \multicolumn{2}{c}{\textbf{Dev}} & \multicolumn{2}{c}{\textbf{Test}}\\ \cline{2-7}
& \#entity & \#sent & \#entity & \#sent & \#entity & \#sent \\ \hline
CoNLL-2003 & $23499$ & $14041$ & $5942$ & $3250$ & $5648$ & $3453$ \\
Taobao & $29397$ & $6000$ & $4941$ & $998$ & $4866$ & $1000$ \\
Youku & $12754$ & $8001$ & $1580$ & $1000$ & $1570$ & $1001$ \\
\hline
\end{tabular}}
\caption{\label{data statistics} Data statistics for CoNLL-2003, Taobao and Youku. `\#entity' represents the number of entities, and `\#sent' is the number of sentences.}
\vspace{-3pt}
\end{table}

\vspace{-3pt}
\subsection{Annealing Technique for $\lambda$} 
Intuitively, the top-$K$ paths decoded by the algorithm could be of poor quality at the beginning of training, because the model's parameters used in decoding haven't been trained adequately. Therefore, we employ an annealing technique to adapt $\lambda$ during training as:
\begin{equation*} \label{lambda annealing}
    \vspace{-3pt}
    \lambda(b) = \exp\left[\gamma\left(\frac{b}{B}-1\right)\right],
\end{equation*}
where $b$ is the current training step, $B$ is the total number of training steps, and $\gamma$ is the hyper-parameter used to control the accelerated speed of $\lambda$. The coefficient $\lambda$ increases rapidly at the latter half of the training, enforcing the model to extracting more information from the top-$K$ paths.

\vspace{-3pt}
\subsection{Iterative Sample Selection} 
Due to the incomplete annotation, there exist some samples whose $q$ distributions are poorly estimated. We use an idea of sample selection to deal with these samples. In each iteration, after training a model $M(\wv_i)$ on $(k-1)$ folds, we utilize $M(\wv_i)$ to decode a most possible path $\hat{\yv}$ for $\xv \in D_i$, and assign a probability score $s=p_{\wv_i}(\hat{\yv}|\xv)$ to $\xv$ at the meantime. Iterative sample selection is to select the samples with probability scores beyond a threshold to construct new training data, which are used in the training phase of $k$-fold cross-validation in the next iteration (more Algorithm details can be found in Algorithm \ref{detail-alg}). We use this strategy to help model identify the gold path effectively with high-quality samples.

\section{Experiments}

\subsection{Dataset}
To benchmark AdaK-NER against its SOTA alternatives in realistic settings, we consider one standard English dataset and two Chinese datasets from Financial Technology Industry:
($i$) {\it CoNLL-2003 English} \cite{sang2003introduction}: annotated by person (PER), location (LOC) and organization (ORG) and miscellaneous (MISC). ($ii$) {\it Taobao\footnote{\url{http://www.taobao.com}}}: a Chinese e-commerce site. The model type (PATTERN), product type (PRODUCT), brand type (BRAND) and the other entities (MISC) are recognized in the dataset. ($iii$) {\it Youku\footnote{\url{http://www.youku.com}}}: a Chinese video-streaming website with videos from various domains. Figure type (FIGURE), program type (PROGRAM) and the others (MISC) are annotated.
Statistics for datasets are presented in Table \ref{data statistics}.

We randomly remove a proportion of entities and all O labels to construct the incomplete annotation, with $\rho$ representing the ratio of entities that keep annotated. We employ two schemes for removing entities:
\begin{itemize}
    \item \textbf{Random-based Scheme} is simply removing entities by random \cite{jie2019better,li2021empirical}, which simulates the situation that a given entity is not recognized by an annotator.
    \item \textbf{Entity-based Scheme} is removing all occurrences of a randomly selected entity until the desired amount remains \cite{mayhew2019named,effland2021partially,wang2019crossweigh}. For example, if the entity ‘Bahrain’ is selected, then every occurrence of ‘Bahrain’ will be removed. This slightly complicated scheme matches the situation that some entities in a special domain could not be recognized by non-expert annotators.
\end{itemize}
According to the low recall of entities tagged by non-speaker annotators in Mayhew~\shortcite{mayhew2019named}, we set $\rho = 0.2$ and $\rho = 0.4$ in our experiments.

\subsection{Experiment Setup}
\paragraph{Evaluation Metrics.}
We consider the following performance metrics: Precision ($P$), Recall ($R$), and balanced F-score ($F_1$). These metrics are calculated based on the true entities and the recognized entities. $F_1$ score is the main metric to evaluate the final NER models of baselines and our approach.

\begin{table*}
\centering
\scalebox{0.88}{\begin{tabular}{clccccccccc}
\hline 
\multirow{2}*{\textbf{Ratio}} &
\multirow{2}*{\textbf{Model}} & \multicolumn{3}{c}{\textbf{CoNLL-2003}} & \multicolumn{3}{c}{\textbf{Taobao}} & \multicolumn{3}{c}{\textbf{Youku}}\\ \cline{3-11}
& & $\mathbf{P\uparrow}$ & $\mathbf{R\uparrow}$ & $\mathbf{F_1\uparrow}$ & $\mathbf{P\uparrow}$ & $\mathbf{R\uparrow}$ & $\mathbf{F_1\uparrow}$ & $\mathbf{P\uparrow}$ & $\mathbf{R\uparrow}$ & $\mathbf{F_1\uparrow}$ \\ \hline
\multirow{4}*{$\rho = 0.2$} &
BERT CRF & $81.42$ & $15.05$ & $25.40$ & $\mathbf{83.11}$ & $24.06$ & $37.32$ & $64.85$ & $20.45$ & $31.09$ \\
& BERT Fuzzy CRF & $17.94$ & $\mathbf{88.14}$ & $29.81$ & $41.48$ & $\mathbf{80.39}$ & $54.72$ & $22.74$ & $\mathbf{84.65}$ & $35.85$ \\
& BERT weighted CRF & $85.03$ & $82.65$ & $83.82$ & $70.06$ & $57.85$ & $63.37$ & $70.18$ & $38.98$ & $50.12$ \\ \cline{2-11}
& AdaK-NER & $\mathbf{87.05}$ & $86.74$ & $\mathbf{86.89}$ & $74.24$ & $78.89$ & $\mathbf{76.50}$ & $\mathbf{78.21}$ & $79.96$ & $\mathbf{78.09}$ \\ \hline
\multirow{4}*{$\rho = 0.4$} &
BERT CRF & $80.07$ & $51.25$ & $62.49$ & $84.76$ & $47.68$ & $61.03$ & $78.89$ & $50.70$ & $61.73$ \\
& BERT Fuzzy CRF & $14.89$ & $86.61$ & $25.41$ & $43.51$ & $85.02$ & $57.56$ & $30.88$ & $84.01$ & $45.16$ \\
& BERT weighted CRF & $85.40$ & $88.69$ & $87.01$ & $73.17$ & $81.09$ & $76.93$ & $74.99$ & $82.29$ & $78.47$ \\ \cline{2-11}
& AdaK-NER & $87.47$ & $88.70$ & $\mathbf{88.08}$ & $74.08$ & $80.13$ & $\mathbf{76.99}$ & $78.38$ & $81.53$ & $\mathbf{79.93}$ \\ \hline
$\rho=1.0$ & BERT CRF & $91.34$ & $92.36$ & $91.85$ & $86.01$ & $88.20$ & $87.09$ & $83.20$ & $84.52$ & $83.85$ \\
\hline
\end{tabular}}
\caption{\label{model comparison} Performance comparison between different models on three datasets with Random-based Scheme.}
\end{table*}

\begin{table*}
\centering
\scalebox{0.88}{\begin{tabular}{clccccccccc}
\hline 
\multirow{2}*{\textbf{Ratio}} &
\multirow{2}*{\textbf{Model}} & \multicolumn{3}{c}{\textbf{CoNLL-2003}} & \multicolumn{3}{c}{\textbf{Taobao}} & \multicolumn{3}{c}{\textbf{Youku}}\\ 
& & $\mathbf{P\uparrow}$ & $\mathbf{R\uparrow}$ & $\mathbf{F_1\uparrow}$ & $\mathbf{P\uparrow}$ & $\mathbf{R\uparrow}$ & $\mathbf{F_1\uparrow}$ & $\mathbf{P\uparrow}$ & $\mathbf{R\uparrow}$ & $\mathbf{F_1\uparrow}$ \\ \hline
\multirow{4}*{$\rho = 0.2$} &
BERT CRF & $\mathbf{86.79}$ & $18.36$ & $30.31$ & $39.62$ & $10.58$ & $16.70$ & $69.10$ & $15.67$ & $25.55$ \\
& BERT Fuzzy CRF & $15.99$ & $\mathbf{86.30}$ & $26.98$ & $42.49$ & $\mathbf{82.33}$ & $56.05$ & $27.79$ & $\mathbf{86.37}$ & $42.05$ \\
& BERT weighted CRF & $83.40$ & $70.96$ & $76.68$ & $\mathbf{73.49}$ & $52.63$ & $61.33$ & $74.71$ & $32.55$ & $45.34$ \\ \cline{2-11}
& AdaK-NER & $86.32$ & $71.72$ & $\mathbf{78.35}$ & $73.24$ & $76.59$ & $\mathbf{74.88}$ & $\mathbf{78.86}$ & $75.54$ & $\mathbf{76.20}$ \\ \hline
\multirow{4}*{$\rho = 0.4$} &
BERT CRF & $\mathbf{86.68}$ & $34.26$ & $49.11$ & $\mathbf{78.43}$ & $39.68$ & $52.70$ & $62.16$ & $35.16$ & $44.91$ \\
& BERT Fuzzy CRF & $13.84$ & $\mathbf{84.60}$ & $23.79$ & $42.24$ & $81.07$ & $55.54$ & $32.10$ & $82.87$ & $46.27$ \\
& BERT weighted CRF & $84.68$ & $76.91$ & $80.61$ & $74.65$ & $79.57$ & $77.03$ & $75.67$ & $80.64$ & $78.08$ \\ \cline{2-11}
& AdaK-NER & $85.48$ & $77.85$ & $\mathbf{81.49}$ & $74.58$ & $\mathbf{80.54}$ & $\mathbf{77.44}$ & $\mathbf{79.01}$ & $81.02$ & $\mathbf{80.00}$ \\ \hline
$\rho=1.0$ & BERT CRF & $91.34$ & $92.36$ & $91.85$ & $86.01$ & $88.20$ & $87.09$ & $83.20$ & $84.52$ & $83.85$ \\
\hline
\end{tabular}}
\caption{\label{model comparison:entity} Performance comparison between different models on three datasets with Entity-based Scheme.}
\end{table*}

\paragraph{Baselines.}
We consider several strong baselines to compare with the proposed methods, including BERT with conventional CRF (or CRF for abbreviation) \cite{lafferty2001conditional}, BERT with Fuzzy CRF \cite{shang2018learning}, and BERT with weighted CRF presented by Jie~\shortcite{jie2019better}. CRF regards all unannotated tokens as O label to form complete paths, while Fuzzy CRF treats all possible paths compatible with the incomplete path with equal probability. Weighted CRF assigns an estimated distribution to all possible paths derived from the incomplete path to train the model. 

\paragraph{Training details.}
We employ BERT model \cite{devlin2018bert} as the neural architecture for baselines and our AdaK-NER. Specifically, we use pretrained Chinese BERT with whole word masking \cite{cui2019cross} for the Chinese datasets and pretrained BERT with case-preserving WordPiece \cite{devlin2018bert} for CoNLL-2003 English dataset. 
Unless otherwise specified, we set the hyperparameter over [top $K$] as 5 by default for illustrative purposes. 
Based on the fact that a larger $k$-fold value has a negligible effect \cite{jie2019better}, we choose to split the training data into $2$ folds ({\it i.e.}, $k=2$). 
We initialize $q$ distribution by assign each unannotated token as O label to form complete paths, and iteratively updated $q$ by k-fold cross-validation stacking.
Empirically, we set the iteration number to 10, which is enough for our model to converge. 


\vspace{-3pt}
\subsection{Experimental Results}
\vspace{-1pt}
To validate the utility of our model, we consider a wide range of real-world tasks experimentally with entity keeping ratio $\rho=0.2$ and $\rho=0.4$. We present the results with Random-based Scheme in Table \ref{model comparison} and Entity-based Scheme in Table \ref{model comparison:entity}. 
We compare the performance of our method to other competing solutions, with each baseline carefully tuned to ensure fairness.  
In all cases, CRF has high precision and low recall because it labels all the unannotated tokens as O label. 
In contrast, taking all possible paths into account yields the mismatch of the gold path, hence Fuzzy CRF recalls more entities. 
Weighted CRF outperforms CRF and Fuzzy CRF, indicating that distribution $q$ should be highly skewed rather than uniformly distributed. 

With adaptive $K$-best loss, candidate mask, annealing technique and iterative sample selection approach, our approach AdaK-NER performs strongly, exhibits high precision and high recall on all datasets and gives best results in $F_1$ score over the other three models. The improvement is especially remarkable on Chinese Taobao and Youku datesets for $\rho=0.2$, as it delivers over $13\%$ and $27\%$ increase in $F_1$ score with Random-based Scheme, while over $13\%$ and $30\%$ increase with Entity-based Scheme.

Note that in CoNLL-2003 and Youku, the $F_1$ score of AdaK-NER with Random-based Scheme is only roughly $5\%$ lower than that of CRF trained on complete data ($\rho = 1$), while we build AdaK-NER on the training data with only $20\%$ entities available ($\rho = 0.2$). In the other Chinese dataset, our model also achieves encouraging improvement compared to the other methods and presents a step toward more accurate incomplete named entity recognition.

Entity-based Scheme is more restrictive, however, our model still achieves best $F_1$ score compared with other methods.
The overall results show AdaK-NER achieves state-of-the-art performance compared to various baselines on both English and Chinese datasets with incomplete annotations.

\begin{figure}[t!]
\centering
\hspace{-7em}
\includegraphics[width=.25\textwidth]{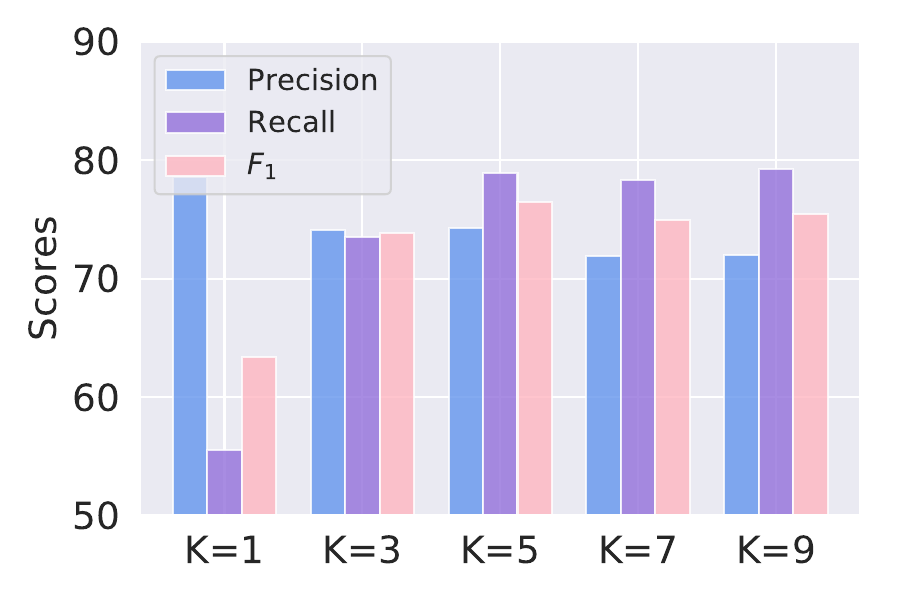}
\hspace{-1em}
\includegraphics[width=.24\textwidth]{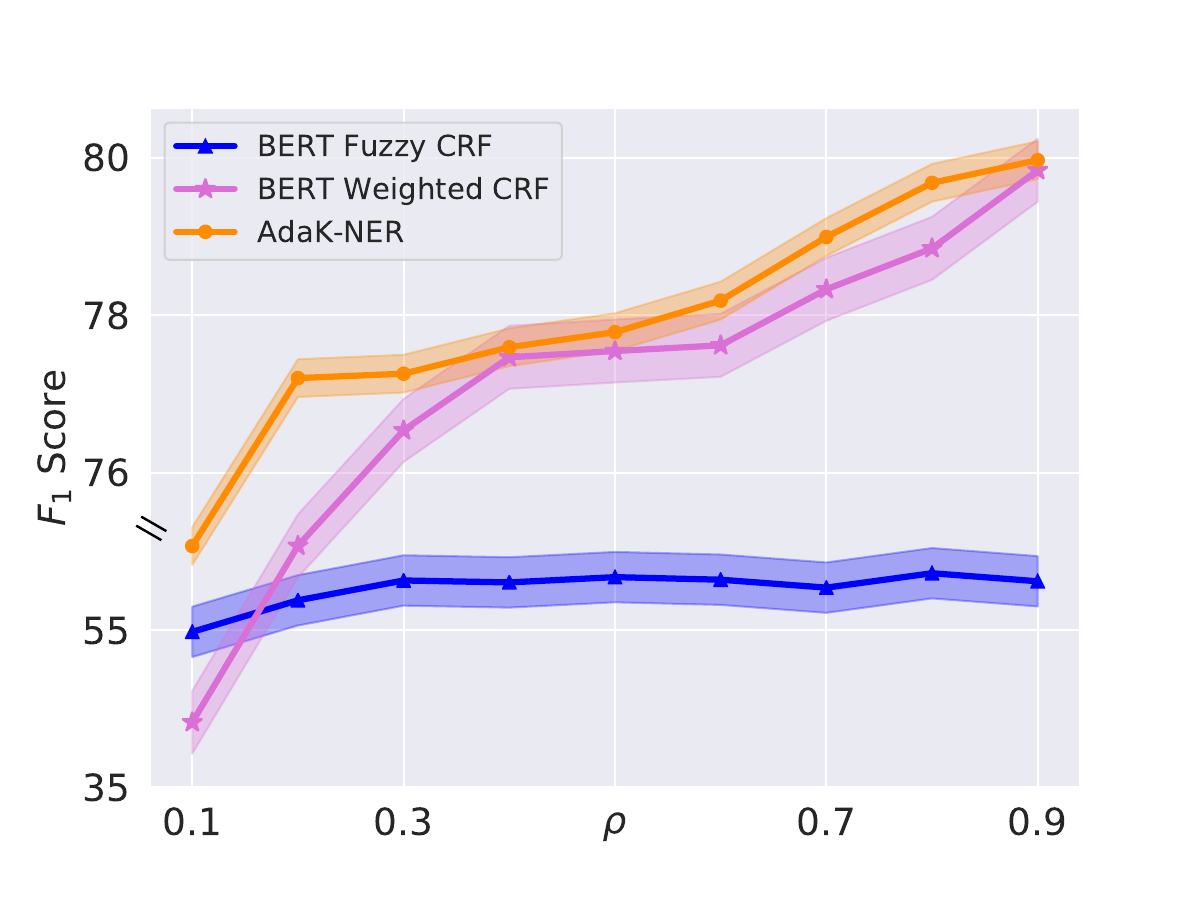}
\hspace{-7em}
\vspace{-.6em}
\caption{(left) Sensitivity analysis of top truncation K. A smaller K is more sensitive to the truncation. (right) $F_1$ score comparison between Fuzzy CRF, weighted CRF and our model on Taobao dataset across different $\rho$ selection with Random-based Scheme}
\label{k comparison}
\end{figure}

\begin{table*}
\centering
\scalebox{0.97}
{\begin{tabular}{lccccccccc}
\hline 
\multirow{2}*{\textbf{Model}} & \multicolumn{3}{c}{\textbf{CoNLL-2003}} & \multicolumn{3}{c}{\textbf{Taobao}} & \multicolumn{3}{c}{\textbf{Youku}}\\ \cline{2-10}
& $\mathbf{P\uparrow}$ & $\mathbf{R\uparrow}$ & $\mathbf{F_1\uparrow}$ & $\mathbf{P\uparrow}$ & $\mathbf{R\uparrow}$ & $\mathbf{F_1\uparrow}$ & $\mathbf{P\uparrow}$ & $\mathbf{R\uparrow}$ & $\mathbf{F_1\uparrow}$ \\ \hline
w/o $K$-best loss & $88.55$ & $80.49$ & $84.33$ & $78.64$ & $53.12$ & $63.41$ & $62.26$ & $39.3$ & $48.18$ \\
w/o weighted loss & $83.79$ & $82.84$ & $83.32$ & $47.85$ & $66.15$ & $55.53$ & $73.46$ & $71.59$ & $72.52$ \\
w/o annealing & $88.36$ & $84.01$ & $86.13$ & $76.09$ & $60.05$ & $67.13$ & $80.98$ & $62.29$ & $70.79$ \\
w/o $K$-best candidates & $84.42$ & $73.76$ & $78.73$ & $72.75$ & $56.62$ & $63.68$ & $73.94$ & $58.03$ & $65.02$ \\
w/o self-built candidates & $87.52$ & $86.33$ & $\mathbf{86.92}$ & $72.38$ & $77.44$ & $74.82$ & $77.88$ & $74.46$ & $76.13$ \\
w/o candidate mask & $84.97$ & $86.51$ & $85.73$ & $68.29$ & $79.16$ & $73.32$ & $73.40$ & $79.81$ & $76.47$ \\
w/o sample selection & $86.64$ & $86.03$ & $86.33$ & $72.88$ & $79.59$ & $76.09$ & $78.48$ & $79.43$ & $\mathbf{78.95}$ \\ \hline
AdaK-NER & $87.05$ & $86.74$ & $\mathbf{86.89}$ & $74.24$ & $78.89$ & $\mathbf{76.50}$ & $78.21$ & $79.96$ & $78.09$ \\
\hline
\end{tabular}}
\caption{\label{ablation study} Ablation study for AdaK-NER on three datasets with Random-based Scheme for $\rho=0.2$.}
\end{table*}

\vspace{-.1em}
\paragraph{The Effect of $K$.}
As discussed in Section \ref{section_4_2} and \ref{section_4_1}, the parameter $K$ can affect the learning procedure from two aspects. We compare the performance of different $K$ on Taobao dataset with Random-based Scheme and $\rho = 0.2$. The hyperparameter over [top $K$] is selected from $\{1, 3, 5, 7, 9\}$ on the validation set. As illustrated in Figure \ref{k comparison}, a relatively large $K$ delivers better empirical results, and the metrics (precision, recall and $F_1$) are pretty close when $K=5, 7, 9$. Meanwhile, a smaller $K$ can narrow down the possible paths more effectively in theory. Hence we favor $K=5$ which might be a balanced choice.

\paragraph{The Effect of $\rho$.}
We further examine annotation rate ($\rho$) interacts with learning. We plot $F_1$ score on Taobao dataset with Random-based Scheme across varying annotation rate in Figure \ref{k comparison}. 
The annotation removed with large $\rho$ inherits the annotation removed with the small $\rho$. All the performance deliver better results with the increase of $\rho$. Our model consistently outperforms weighted CRF and Fuzzy CRF, and the improvement is significant when $\rho$ is relatively small, which indicates our model is especially powerful when the annotated tokens are fairly sparse. 

\vspace{-3.5pt}
\subsection{Ablation Study}
\vspace{-2pt}
To investigate the effectiveness of the proposed strategies used in AdaK-NER, we conduct the following ablation with Random-based Scheme and $\rho = 0.2$. As shown in Table \ref{ablation study}, the adaptive $K$-best loss contributes most to our model on the three datasets. It helps our model achieve higher recall while preserving acceptable precision. Especially on Youku dataset, removing it would cause a significant drop on recall by $40\%$. The weighted CRF loss is indispensable, and annealing method could help our model achieve better results.
Candidate mask is attributed to promote precision while keeping high recall. Both $K$-best candidates and self-built candidates facilitate the model performance. 
Iterative sample selection makes a positive contribution to 
our model on CoNLL-2003 and Taobao, whereas it slightly hurts the performance on Youku.
In general, incorporating these techniques enhances model performance on incomplete annotated data.

\vspace{-5pt}
\section{Related Works}
\vspace{-3pt}

\paragraph{Pre-trained Language Models} has been an emerging direction in NLP since Google launched BERT \cite{devlin2018bert} in 2018. With the powerful Transformer architecture, several pre-trained models, such as BERT and generative pre-training model (GPT), and their variants have achieved state-of-the-art performance in various NLP tasks including NER \cite{devlin2018bert,liu2019roberta}.
Yang~\shortcite{yang2019xlnet} proposed a pre-trained permutation language model (XLNet) to overcome the limitations of denoising autoencoding based pre-training. Liu~\shortcite{liu2019roberta} demonstrated that more data and more parameter tuning could benefit pre-trained language models, and released a new pre-trained model (RoBERTa). 
To follow the trend, we use BERT as our neural model in this work.

\paragraph{Statistical Modeling} has been widely employed in sequence labeling. 
Classical models learn label sequences through graph-based representation, with prominent examples such as Hidden Markov Model (HMM), Maximum Entropy Markov Models (MEMM) and Conditional Random Fields (CRF) \cite{lafferty2001conditional}. Among them, CRF is an optimal model, since it resolves the labeling bias issue in MEMM and doesn't require the unreasonable independence assumptions in HMM. 
However, conventional CRF is not directly applicable to the incomplete annotation situation. Ni~\shortcite{ni2017weakly} select the sentences with the highest confidence, and regarding missing labels as O. Another line of work is to replace CRF with Partial CRF \cite{nooralahzadeh2019reinforcement,huang2019learning} or Fuzzy CRF \cite{shang2018learning}, which assign unlabeled words with all possible labels and maximize the total probability \cite{yang2018distantly}. Although these works have led to many promising results, they still need external knowledge for high-quality performance. Jie~\shortcite{jie2019better} presented a weighted CRF model which is most closely related to our work. They estimated a proper distribution for all possible paths derived from the incomplete annotations. Our work enhances Fuzzy CRF by reducing the possible paths by a large margin, aiming to better focus on the gold path.


\vspace{-.7em}
\section{Conclusion}
\vspace{-3pt}

In this paper, we explore how to build an effective NER model by only using incomplete annotations. We propose two major strategies including introducing a novel adaptive $K$-best loss and a mask based on $K$-best candidates and self-built candidates to help our model better focus on the gold path. 
The results show that our approaches can significantly improve the performance of NER model with incomplete annotations.

\clearpage

\bibliographystyle{named}
\bibliography{ijcai22}

\end{document}


\appendix
\renewcommand{\thetable}{S\arabic{table}}
\renewcommand{\thefigure}{S\arabic{figure}}
\renewcommand{\thealgorithm}{S\arabic{algorithm}}

\section{Dataset}
\label{appendixA}
\vspace{-.3em}
Our model has been applied not only on CoNLL-2003 \cite{sang2003introduction}, but also on two Chinese datasets ({\it i.e.} Taobao and Youku\footnote{http://www.taobao.com and http://www.youku.com}). CoNLL-2003 is a well-accepted NER English benchmark. Taobao and Youku are two industry datasets from Financial Technology, crawled and manually annotated by Jie~\shortcite{jie2019better}. We introduce the details of thses two Chinese datasets in this section. 
Taobao and Youku datasets contain product and video titles in Chinese. There are 51\% and 41.7\% of entity labels per sentence in Taobao and Youku separately.

\paragraph{Taobao} is an e-commerce site. Jie~\shortcite{jie2019better} annotate the data of product titles from the website with 4 categorized types as well as 9 fine-grained entity types, including PATTERN (Model Type), PRODUCT (Product Description, Core Product), BRAND (Brand Description, Core Brand) and MISC (Location, Person, Literature, Product Specification). We utilize 4 categorized types in our experiments.

\paragraph{Youku} is a video-streaming website with videos from various domains. Jie~\shortcite{jie2019better} crawl the video titles from the website and mark the entities with 3 categorized types and 9 fine-grained entity types, including FIGURE (Figure Type), PROGRAM (Variety Show, Movie, Animation, TV Drama) and MISC (Character, Number, Location, Song). We use 3 categorized types here.

\section{Approach Details}
\label{appendixB}

\begin{figure*}[!htbp]
\centering
\includegraphics[width=.9\textwidth]{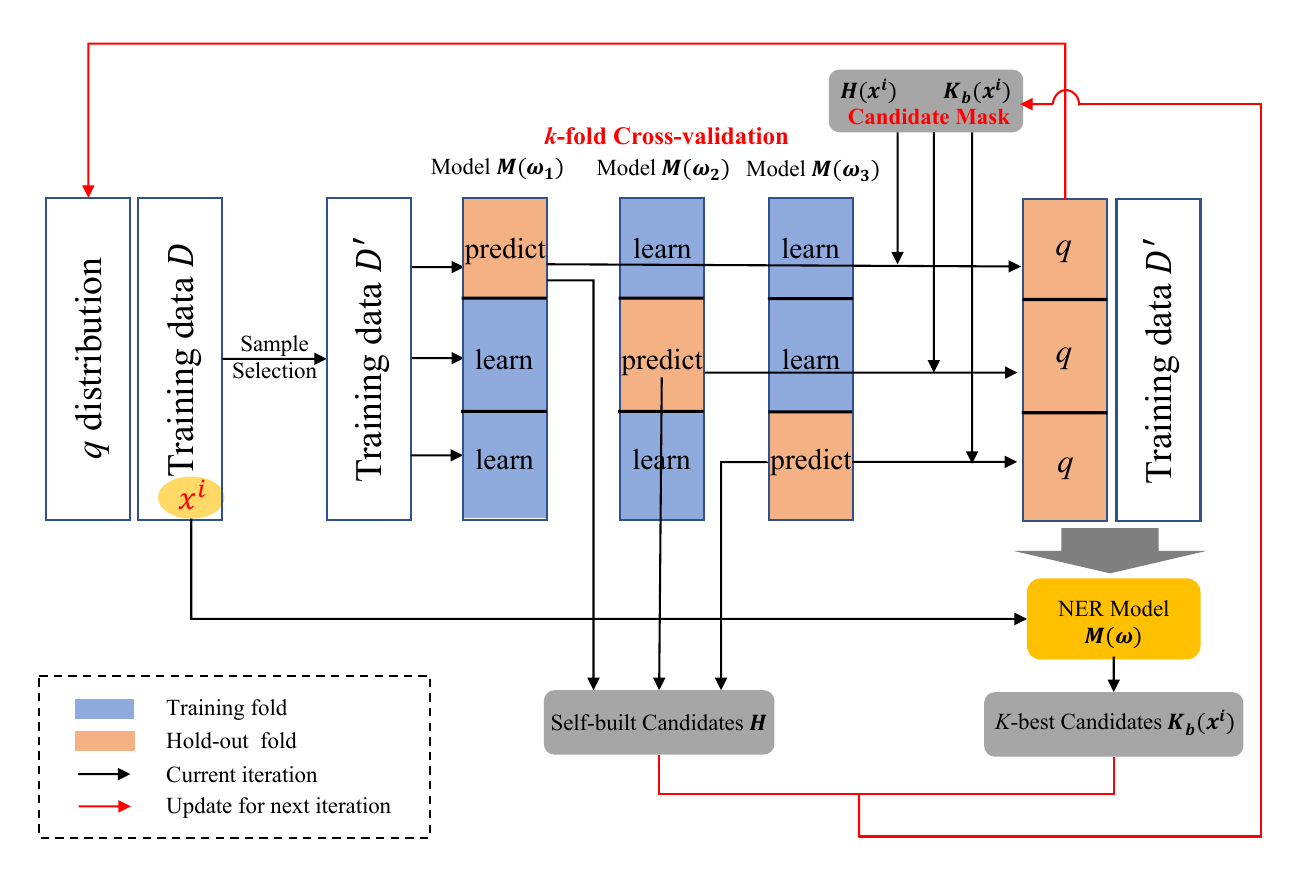}
\caption{The flow chart of AdaK-NER algorithm. After completing all iterations, we could gain the final NER model $M(\wv)$ which is trained on all training data.}
\label{algo_procedure}
\end{figure*}


Jie~\shortcite{jie2019better} presented two different ways to define $q(\yv|\xv)$ including the hard mode and the soft mode. The hard mode obtains a collapsed distribution which regards the probability of a single path as 1, while the soft mode assigns a probability to each possible path. We further propose an interpolation mode (softmax mode) to bridge the hard and soft mode, adjust $q$ from soft mode to increase the probabilities for those with high confidence and decrease for others.

In the hard mode, after training model $M(\wv_i)$ on $(k-1)$ folds $\{D_1, \cdots, D_{i-1}, D_{i+1}, \cdots, D_k\}$, a constrained Viterbi procedure\footnote{The constrained decoding ensures the resulting complete path is a possible label sequence compatible with the incomplete annotations.} is used to decode a single path for each sentence in the hold-out fold. 
For all $\xv\in D_i$ (the hold-out fold), $\hat{\yv} = \arg\max_{\yv}p_{\wv_i}(\yv|\xv)$, and
\begin{align*}
q_{\wv_i}(\yv|\xv) =\begin{cases}
    1,~~ \yv=\hat{\yv},\\
    0,~~ \yv\neq \hat{\yv}.
    \end{cases}\end{align*}

In the soft mode, we utilize a constrained forward-backward procedure \cite{jie2019better} to compute the marginal probabilities associated with each label at each unannotated token. The product of all marginal probabilities in a complete path is set to the probability of the path.
For all $\xv\in D_i,q(\yv|\xv) = p_{\wv_i}(\yv|\xv)$ where $\yv \in C(\yv_u)$.

In the interpolation mode, the probability of each complete label sequence is a temperature softmax of $\log p_{\wv_i}$: $q_{\wv_i}(\yv|\xv)= \mbox{softmax}(p_{\wv_i}(\yv|\xv), T)$ (equation \ref{temperature} in main text). One reasonable scheme for choosing $T$ would be starting from a relative high temperature and gradually lower down in order to explore more early on and exploit the resulting knowledge later. Because the results of the hard and soft approach don't differ much \cite{jie2019better}, we choose to use the hard approach in our experiments for simplicity. 

Note that the estimation of $q$ depends on its initialization (we discuss the initialization in Supplementary Material (SM) \ref{appendixC}), we iterate the $k$-fold cross-validation procedure until $q$ converges.

\vspace{-.3em}
\section{Experiment Details}
\label{appendixC}
\vspace{-.3em}
We introduce the detailed settings of our experiments in this section.

\vspace{-.3em}
\paragraph{Training details.}
We employ BERT model \cite{devlin2018bert} as the neural architecture for baselines and our AdaK-NER. Specifically, we use pretrained Chinese BERT with whole word masking \cite{cui2019cross} for the Chinese datasets and pretrained BERT with case-preserving WordPiece \cite{devlin2018bert} for CoNLL-2003 English dataset. Our model is trained on Adam optimizer \cite{kingma2014adam} with a batch size of $32$ for all the parameters. We use a value $1e-4$ as the learning rate for the Chinese datasets and $2e-4$ for the English dataset. Our model is trained with $10$ epochs. The max allowed sequence length is set to $180$ for the Chinese datasets and $256$ for the English dataset. The hyperparameter over [top-$K$] is selected from $\{1, 3, 5, 7, 9\}$ on the validation set, and $5$ is found to deliver the best empirical results. Unless otherwise specified, we set the hyperparameter over [top $K$] as 5 by default for illustrative purposes. Based on the fact that a larger $k$-fold value has a negligible effect, we choose to split the training data into $2$ folds ({\it i.e.}, $k=2$). 
For annealing technique, we set $\gamma$ to $3$ in our experiments. 
For defining $q$ distribution, we set $T\rightarrow 0$ in our $k$-fold cross-validation procedure. Empirically, we set the iteration number to 10, which is enough for our model to converge. All experiments are implemented with PyTorch and executed on a single NVIDIA P100 GPU.

\vspace{-.3em}
\paragraph{Implementation Details.}
Initialization plays an important role in our experiments. Due to our hard mode setting, we initialize $q$ distribution by assign each unannotated token as O label to form complete paths, and iteratively updated $q$ by k-fold cross-validation stacking. For the initialization of the NER model $M$ with parameters $\hat{\wv}$, we could utilize the model trained on 2 folds ($k=2$ in our experiments) for making candidate mask at the very beginning. Thus, in the first iteration we use the model trained on one fold to predict $K_b(\xv)$ for the samples in the same fold and make masks. In the other iterations, we use the model $M(\hat{\wv})$ saved in last iteration to decode $K_b(\xv)$.

\clearpage
\bibliographystyle{named}
\bibliography{ijcai22}